\pdfoutput=1
\documentclass[letterpaper, 10 pt, conference]{ieeeconf} 
\IEEEoverridecommandlockouts                              

\usepackage{graphics} 
\usepackage{epsfig} 
\usepackage{times} 
\usepackage{amsmath} 
\usepackage{amssymb}  
\usepackage{soul,color}
\usepackage{graphicx}

\usepackage{comment}
\usepackage{algorithmic, algorithm} 
\usepackage{fourier} 
\usepackage{dblfloatfix} 
\usepackage[caption=false,font=footnotesize]{subfig} 

\def\mf{\mathbf}
\def\mb{\mathbb}

\def\beq{\begin{equation*}}
\def\eeq{\end{equation*}}
\def\bql{\begin{equation}}
\def\eql{\end{equation}}
\def\bqn{\begin{eqnarray*}}
\def\eqn{\end{eqnarray*}}
\def\bnl{\begin{eqnarray}}
\def\enl{\end{eqnarray}}
\def\bna{\bql\begin{array}{rcl}}
\def\ena{\end{array}\eql}
\def\bnn{\beq\begin{array}{rcl}}
\def\enn{\end{array}\eeq}
\def\bma{\begin{bmatrix}}
\def\ema{\end{bmatrix}}
\def\bmx{\begin{matrix}}
\def\emx{\end{matrix}}
\def\ben{\begin{enumerate}}
\def\een{\end{enumerate}}
\def\bit{\begin{itemize}}
\def\eit{\end{itemize}}
\def\bei{\begin{itemize}}
\def\eei{\end{itemize}}
\def\bet{\begin{tabular}}
\def\eet{\end{tabular}}

\newcommand{\allcaps}[1]{\uppercase\expandafter{#1}}

\def\bfs{\begin{footnotesize}}
\def\efs{\end{footnotesize}}
\def\bss{\begin{small}}
\def\ess{\end{small}}

\def\zd{\mf z_D}
\def\za{\mf z_A}

\title{\LARGE \bf
Vision-based Perimeter Defense via Multiview Pose Estimation
}

\author{Elijah S. Lee$^{1}$, Giuseppe Loianno$^{2}$, Dinesh Jayaraman$^{1}$, and Vijay Kumar$^{1}$
\thanks{We gratefully acknowledge the support from 
ARL DCIST CRA under Grant W911NF-17-2-0181,
NSF under Grants CCR-2112665, CNS-1446592, and EEC-1941529,
ONR under Grants N00014-20-1-2822 and N00014-20-S-B001, 
Qualcomm Research, 
NVIDIA,
Lockheed Martin,
and C-BRIC, a Semiconductor Research Corporation Joint University Microelectronics Program program cosponsored by DARPA.
}
\thanks{$^{1}$The authors are with the GRASP Lab, University of Pennsylvania, Philadelphia, PA 19104, USA. 
{\tt\footnotesize \{elslee, dineshj, kumar\}@seas.upenn.edu}
}%
\thanks{$^{2}$ The author is with the New York University, Tandon School of Engineering, Brooklyn, NY 11201, USA. {\tt\footnotesize email: {loiannog}@nyu.edu.}
}%
}

\begin{document}

\maketitle
\thispagestyle{empty}
\pagestyle{empty}

\begin{abstract}
Previous studies in the perimeter defense game have largely focused on the fully observable setting where the true player states are known to all players. However, this is unrealistic for practical implementation since defenders may have to perceive the intruders and estimate their states. In this work, we study the perimeter defense game in a photo-realistic simulator and the real world, requiring defenders to estimate intruder states from vision. We train a deep machine learning-based system for intruder pose detection with domain randomization that aggregates multiple views to reduce state estimation errors and adapt the defensive strategy to account for this. We newly introduce performance metrics to evaluate the vision-based perimeter defense. Through extensive experiments, we show that our approach improves state estimation, and eventually, perimeter defense performance in both 1-defender-vs-1-intruder games, and 2-defenders-vs-1-intruder games.
\end{abstract}

\section{Introduction}
\begin{figure}[!t]
\centering
\includegraphics[width=7.8cm]{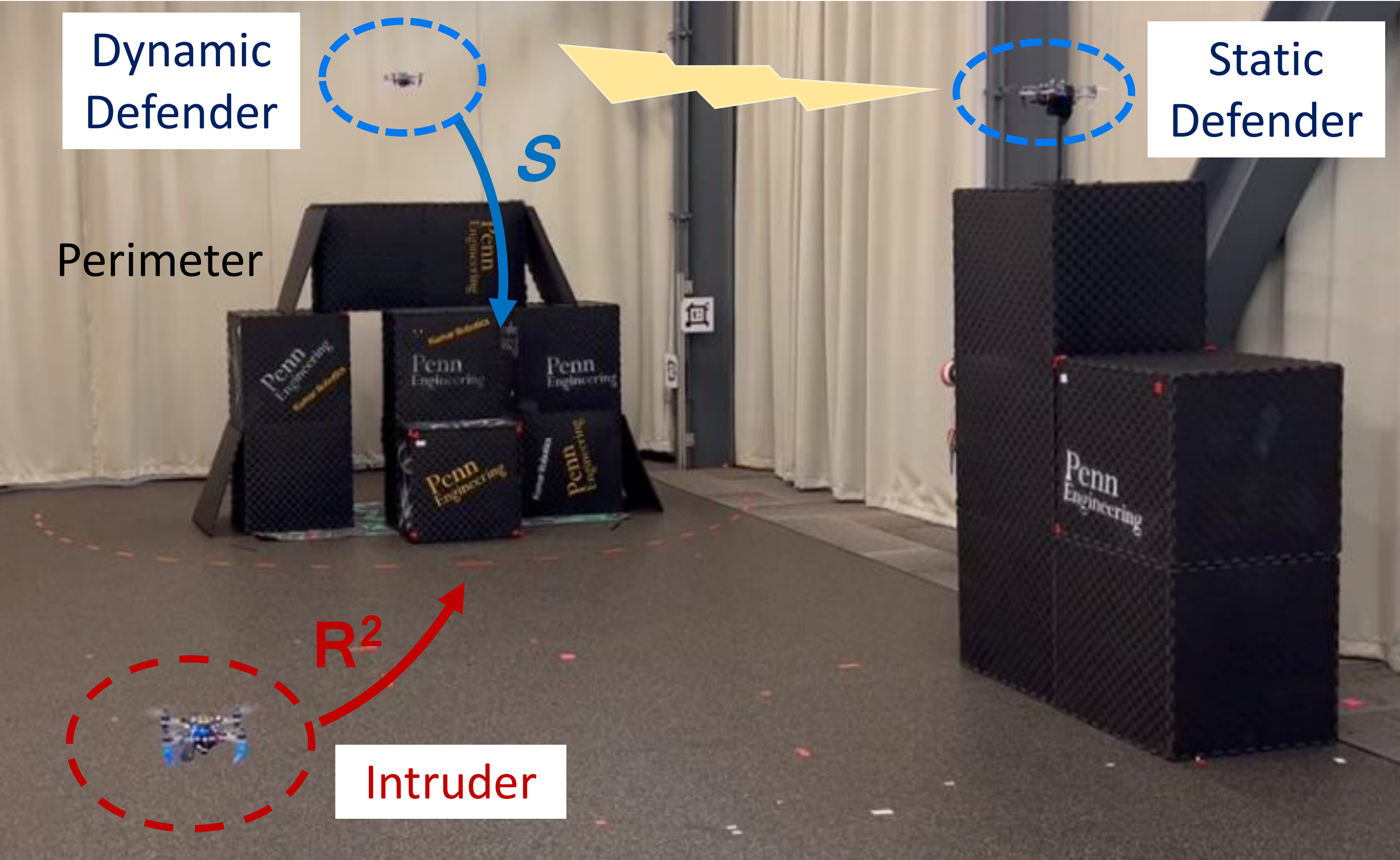}
\caption{
Dynamic defender aims to capture the intruder approaching the perimeter with the help of static defender that allows multiview pose estimation.}
\label{fig:intro}
\end{figure}

In a perimeter defense game, defenders aim to intercept intruders by moving along the perimeter while intruders try to reach the perimeter without being captured by defenders \cite{shishika2020review}. As shown in Fig. \ref{fig:intro}, we study Unmanned Aerial Vehicles (UAVs) equipped with cameras as defenders. A dynamic defender aims to capture an intruder based on its perception while a static defender enables multiview perception. Vision-based UAVs have been well studied in various settings such as power plant \cite{lee2020experimental}, penstock \cite{nguyen2019mavnet}, forest \cite{chen2020sloam}, or disaster sites \cite{lee2016drone}, and all these settings can be real-world use-cases for perimeter defense; for instance, an intruder tries to attack a military base in the forest and a defender aims to capture the intruder.

In large, there are three modules in end-to-end perimeter defense: perception, planning, and control. In this work, the planning module is theoretically derived from our previous work \cite{lee2020perimeter} that the defenders aim to move to the optimal breaching point to maximize their probability to win the game. Since the optimality of the game is given as a Nash equilibrium, we executed the controller to follow the trajectories from the planner.

A number of previous works solve the perimeter defense problem assuming perfect state estimation \cite{lee2020perimeter, shishika2020cooperative, yan2019construction}; however, this is unrealistic for real-world implementations of perimeter defense because defenders may have to perceive the intruders and estimate their states (i.e. positions and velocities) to execute the optimal defense strategies. We newly add a vision-based perception module to predict intruder pose based on domain randomization \cite{tobin2017domain} that runs a lightweight network for scalability.

Our work introduces performance metrics to evaluate the vision-based defense and demonstrates that combining camera views from multiple defenders improves defense performance. In short, our main contributions are:
\begin{itemize}
    \item Realization of vision-based perimeter defense. To the best of our knowledge, we are the first to relax the assumption that all states are known to all players and instead employ vision sensors to acquire necessary state information in a perimeter defense game.
    
    \item Improvement of perimeter defense algorithm performance by incorporating pose estimation by aggregating multiple views with one dynamic and one static defender.
\end{itemize}

\section{Related Work \label{sec:related}}

\subsection{Perimeter defense}
In the perimeter defense game, defenders are tasked to protect a region from intruders. The intruder aims to reach the target without being captured by the defender, while the defender tries to capture the intruder. We refer to \cite{shishika2020review} for a detailed survey. In particular, a cooperative multiplayer perimeter-defense game is solved on a planar game space in \cite{shishika2020cooperative}, and Yan et al.~\cite{yan2019construction} solve a reach-avoid differential game in three-dimensional space. Optimal strategies between aerial defender and ground intruder are obtained in \cite{lee2020perimeter}. All these works assume complete and perfect state estimation, i.e., states are known to all agents. 

Previous works \cite{vidal2002probabilistic, arola2019uav} relax this assumption in pursuit-evasion game. Vidal et al. \cite{vidal2002probabilistic} implement the pursuit-evasion game on real UAVs and UGVs with sensor models for evader detection. Vision-based pursuit-evasion using deep learning is implemented in \cite{arola2019uav}. This work does not assume perfect state estimation in the perimeter defense game and uses vision for state estimation.

\subsection{Pose estimation}
Pose estimation has been studied extensively based on features, and recent methods based on supervised learning 
have been explored. Cosypose \cite{labbe2020cosypose} recovers the 6D pose of objects captured by a set of input images with unknown camera viewpoints by a neural net to recover camera poses where feature-based registration fails. MSL-RAPTOR \cite{ramtoula2020msl} combines learning with a tracker for onboard robotic perception. GDR-Net \cite{wang2021gdr} performs a geometry-guided direct regression network to learn the 6D pose in an end-to-end manner from dense correspondence-based intermediate geometric representations. To avoid extensive labeling in supervised learning methods, domain randomization \cite{tobin2017domain,ren2019domain} generates its own training data in simulation, which learns a robust policy that transfers well to the real world. Our work employs multiview perception that estimates the intruder pose with the network based on domain randomization to improve the defense performance. We focus on training lightweight networks to enable scalable platforms with reasonable accuracy.

\section{Problem Formulation} \label{sec:problem}
We consider a hemisphere defense game where the defender operates on the surface $S$ of the hemisphere while the intruder approaches on the ground plane $\mb R^2$. For instance, a perimeter of a building that defenders aim to protect can be enclosed by a hemisphere. Since defenders cannot pass through the building, they are employed to move along the surface of the dome, which leads to the hemisphere perimeter defense game.

Although there exist optimal strategies for both players in theory \cite{lee2020perimeter}, it can be inferred 
that the theoretical model (e.g. perfect state estimation) may not be realistic for a real-world implementation \cite{lee2021defending}. In this work, we extend our previous work \cite{lee2020perimeter} to incorporate the perception module in the hemisphere perimeter defense game to bridge the gap between theory and practice. This work differs from the previous work in that the dynamics of the UAVs, uncertainty in state estimation, and multiple viewpoints of defenders are newly introduced. Furthermore, we realize the vision-based perimeter defense and test the proposed system in hardware experiments. We also improve vision-based perimeter defense performance by incorporating pose estimation by aggregating multiple views. This section addresses the problem formulation relevant to the vision-based defense.

\subsection{The hemisphere coordinate system}
\begin{figure}[b]
    \centering
    \subfloat[]{\includegraphics[width=0.22\textwidth]{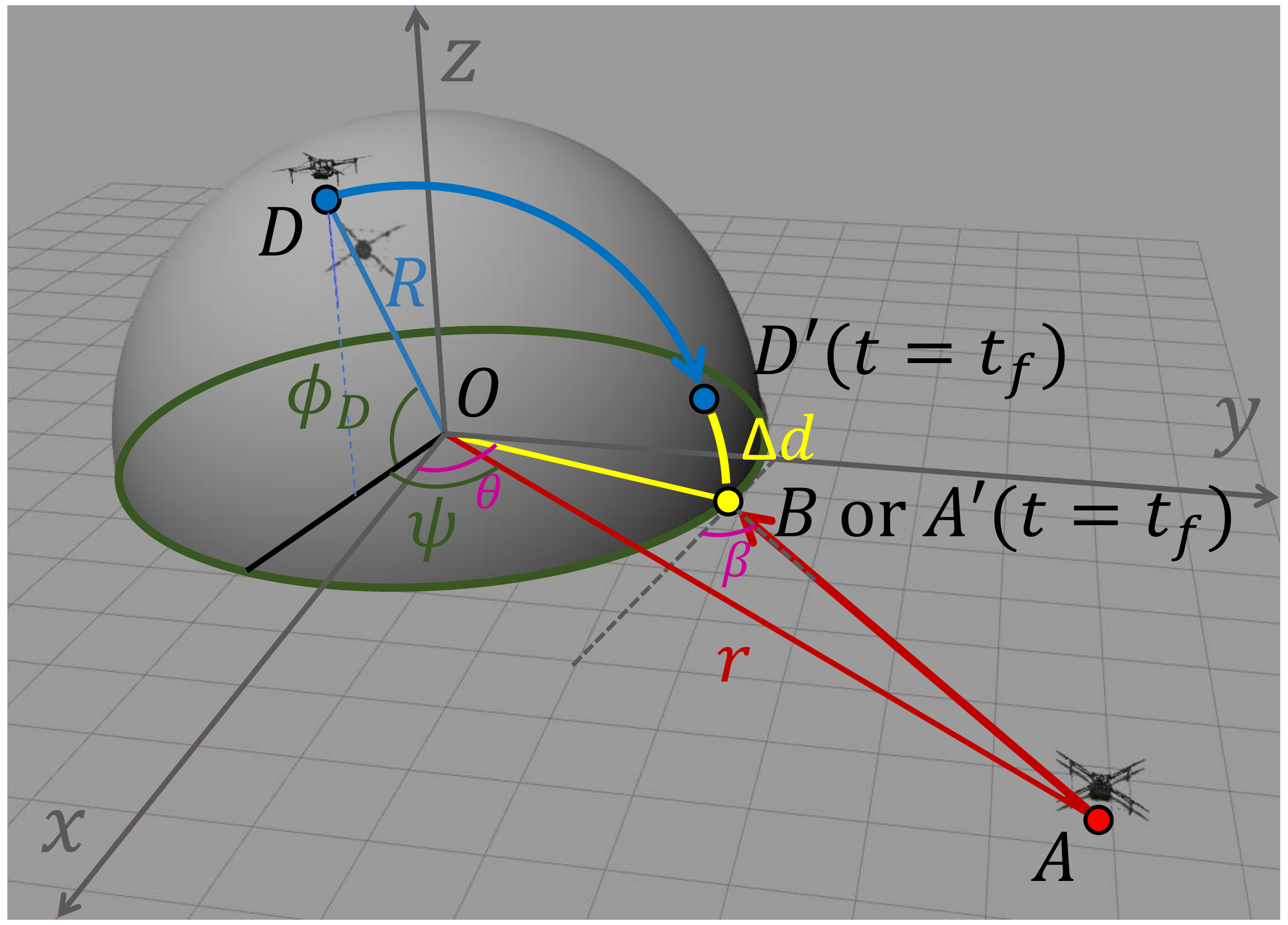}}
    \subfloat[]{\includegraphics[width=0.22\textwidth]{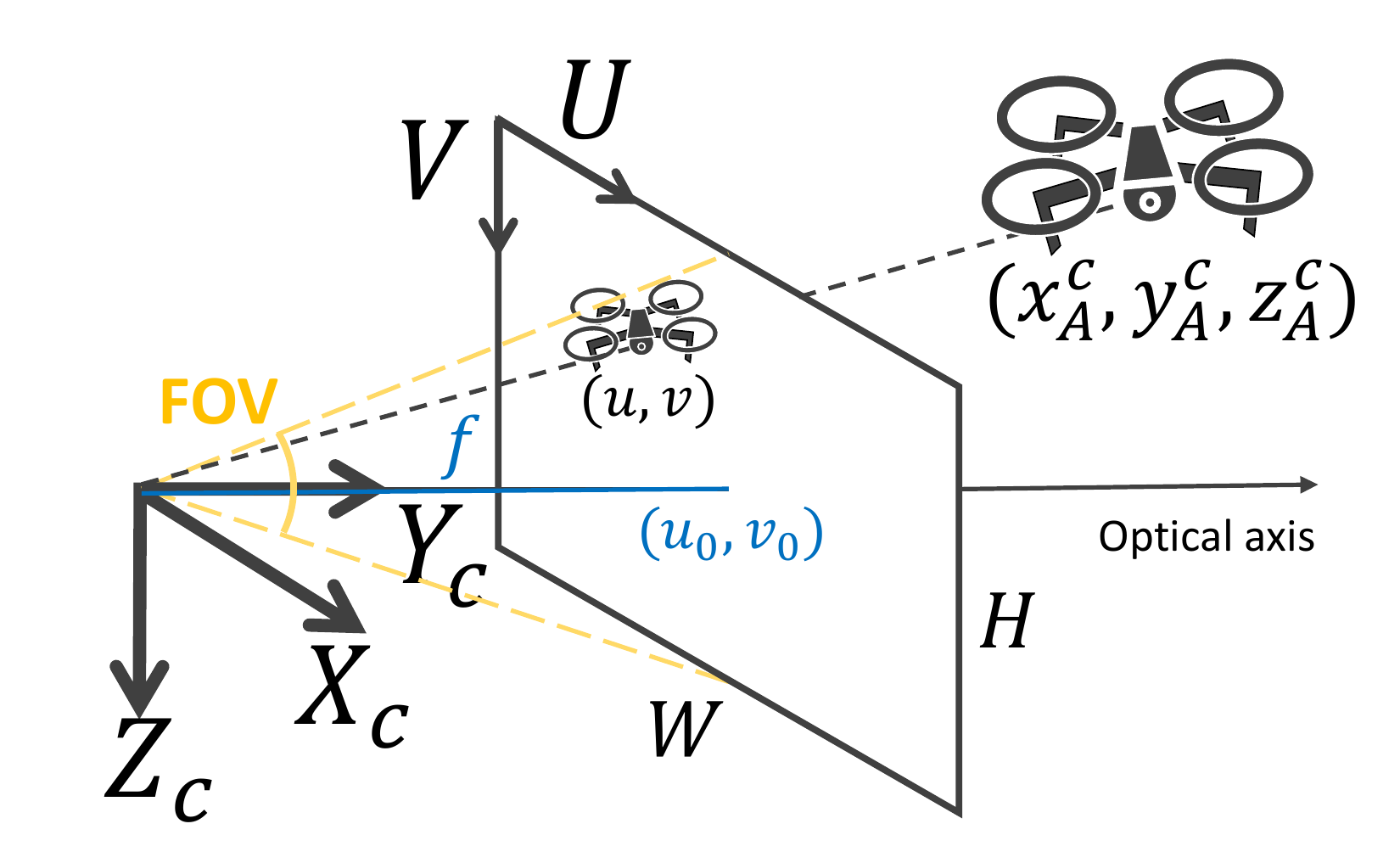}}
    \caption{Coordinates and relevant variables for vision-based hemisphere defense. (a) Hemisphere coordinate system. (b) Camera and image coordinate systems.}
    \label{fig:coord}
\end{figure}

Fig.~\ref{fig:coord}(a) shows the defender $D$, the intruder $A$, and the hemisphere $O$ with radius $R$. The positions of the players in spherical coordinates are: $\mf z_D=[\psi_D,\phi_D,R]$ and $\mf z_A=[\psi_A,0,r]$, where $\psi$ and $\phi$ are the azimuth and elevation angles, which gives the relative position as: $\mf z \triangleq [\psi,\phi,r]$, where $\psi\triangleq \psi_A-\psi_D$ and $\phi\triangleq \phi_D$. At a given time, the speed of defender and intruder are $v_D$ and $v_A$, respectively, and the ratio is defined as $\nu = v_A/v_D\leq1$. Given $\zd$, $\za$, we call $\textit{breaching point}$ as a point on the perimeter at which the intruder aims to reach the target, as shown $B$ in Fig.~\ref{fig:coord}(a). It is proved in \cite{lee2020perimeter} that given the current positions of $\mf z_D$ and $\mf z_A$, there exists a unique breaching point such that the optimal strategy for both defender and intruder is to move towards it, known as \textit{optimal breaching point}.

\begin{figure*}[h]
\centering
\includegraphics[width=0.8\textwidth]{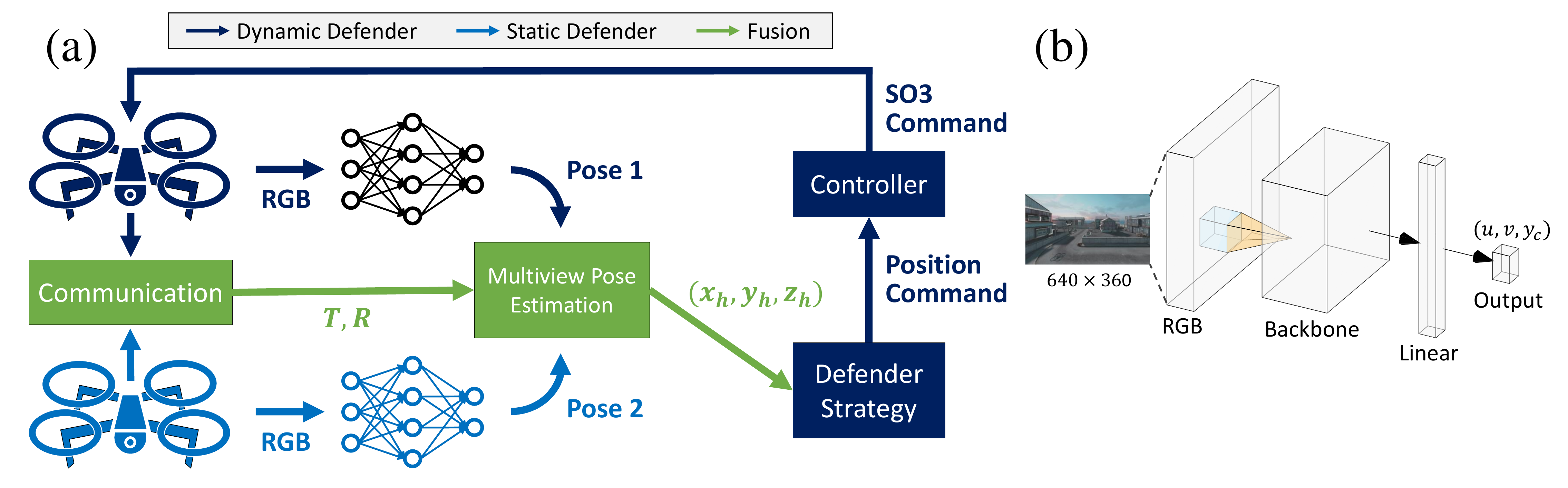}
\caption{(a) Overall framework for vision-based perimeter defense. The dynamic defender may communicate with the static defender to perform multiview pose estimation, and the defender strategy is computed to give position commands to the controller to close the loop. (b) Pose estimation network architecture with ShuffleNet \cite{zhang2018shufflenet} as the backbone. }

\label{fig:overall}
\end{figure*}

\subsection{Intruder pose}
We assume that a camera is mounted on defender to perceive intruder and define the \textit{intruder pose} as $(u,v,y_c)$, which is a 3-tuple combining $u,v$ intruder position in the image coordinates and $y_c$, which indicates $Y_c$ direction in the camera coordinates (i.e., $y_c=y_A^c$), as shown in Fig. \ref{fig:coord}(b). This selection is intuitive in that it is easier to infer the intruder's center position $u,v$ than to infer $(x_A^c,y_A^c,z_A^c)$ from an image, and the $Y_c$ direction is correlated with the known size of the intruder. 
\subsection{Performance metrics}
In theory, we assume perfect state estimation without noise, and thus moving towards the \textit{optimal breaching point} would preserve a Nash equilibrium. However, vision-based perimeter defense may be responsible for uncertainty and noise in estimating intruder pose. Accordingly, we define the performance metrics $\Delta d$ at any time to evaluate the perimeter defense performance as follows:

\bql
\Delta d = 
|\widearc{DB}| - |AB|
\label{eq:deltad}
\eql

If the intruder wins, it reaches the target before the defender does, and $\Delta d$ indicates the geodesic between the defender $D$ and some breaching point $B$. If the defender wins, it is located at some point on the base plane at the end, and $\Delta d$ indicates the negative value of the distance between $D$ and $A$. In both cases, the defender aims to minimize $\Delta d$ while the intruder aims to maximize it. 

\section{Methods \label{sec:method}}
In this work, we build a framework for vision-based perimeter defense by estimating the intruder pose. We propose a pose estimation by aggregating multiple views for perimeter defense with the framework shown in Fig. \ref{fig:overall}. Our motivation is to close the perception-action loop for the dynamic defender to maneuver with a strategy comparable to the optimal strategy from theory, which is moving towards the breaching point. The end goal is to minimize the performance metrics $\Delta d$, expressed in \eqref{eq:deltad}. We aim to show that the multiview perception improves perimeter defense performance.

\subsection{Pose estimation using multiple aggregated views \label{sec:fusion}}
In our defense application, we prepare defenders with two roles: $\textit{dynamic defender}$ and $\textit{static defender}$. These roles enable different capabilities of defenders - dynamic defender operates on a hemisphere surface and aims to capture the intruder by estimating the intruder pose with an on-board camera, and static defender stays in the same position observing the intruder. The multiple views of the scene available from these roles allow multiple data points, which improves the overall perception. We perform an ablation study by combining multiple intruder poses by parameterizing each camera coordinate with weights $\alpha, \delta, \gamma$ (see Section V-D for more details). It is worth noting that the static defender does not try to capture the intruder in the scope of this work, but working with multiple vision-based dynamic defenders will be interesting for future work.

\subsection{Network architecture}
As shown in Fig. \ref{fig:overall}(b), we trained an end-to-end deep neural network for estimating intruder pose. We prefer deep learning over traditional methods for its scalability. The network input is $640\times 360$ image, and the output is $(u,v,y_c)$. We employ ShuffleNet \cite{zhang2018shufflenet} as the backbone for its efficiency, followed by a linear layer to output the pose. The loss function is MSE loss with a weight factor of $\epsilon$ on $u$ and $v$.

\subsection{Perception-action loop}
To close the perception-action loop, the output from multiview pose estimation is given to the dynamic defender's defender strategy module in Fig. \ref{fig:overall}(a), which outputs the position command towards the direction of the optimal breaching point. The SO(3) command \cite{mellinger2011minimum} that consists of thrust and moment to control the robot is then passed to the dynamic defender for control. In this loop, we evaluate the performance metrics $\Delta d$ and aim to show that the proposed approach using multiview perception improves perimeter defense performance.

\section{Experiments \label{sec:experiments}}
In large, we run experiments with a realistic simulator and quadrotor platforms. We first set up the simulator environments (Section V-A) and collect training data (Section V-B) to train the proposed pose estimation networks (Section V-C). Once the parameters are selected by ablation analysis (Section V-D), we further evaluate perimeter defense performance by observing the improvement from the multiview perception in diverse scenarios (Section V-E). Finally, we run hardware experiments by employing quadrotor platforms with on-board cameras to validate that vision-based perimeter defense performance improves with multiview perception (Section V-F).

\subsection{Experimental setup in simulation}
We run the simulation in Flightmare \cite{song2020flightmare}, a modular simulator composed of a rendering engine built on Unity and a physics engine for dynamics simulation. We consider the defense game based on a 1 vs. 1 (a dynamic defender vs. an intruder) and a 2 vs. 1 (a dynamic defender and a static defender vs. an intruder) game. When the game starts, the position of the intruder relative to the dynamic defender is set to desired $\mf z = [\psi,\phi,r]$ in the hemisphere coordinate. The position of the static defender is set to desired $[x_w,y_w,z_w]$ in the world coordinate. The parameters used for the camera and image coordinates are: $W=640, H=360$, FOV$=90, f=180, u_0=320$, and $v_0=180$.

\subsection{Data collection and domain randomization}
Domain randomization \cite{tobin2017domain} utilizes irrelevant variability to learn the relevant features for real-world problems. Practical applications include autonomous driving vehicles \cite{muller2018driving}, drone racing \cite{loquercio2019deep}, and robot manipulation \cite{ren2019domain}.

\begin{figure}[t]
\centering
\includegraphics[width=.47\textwidth]{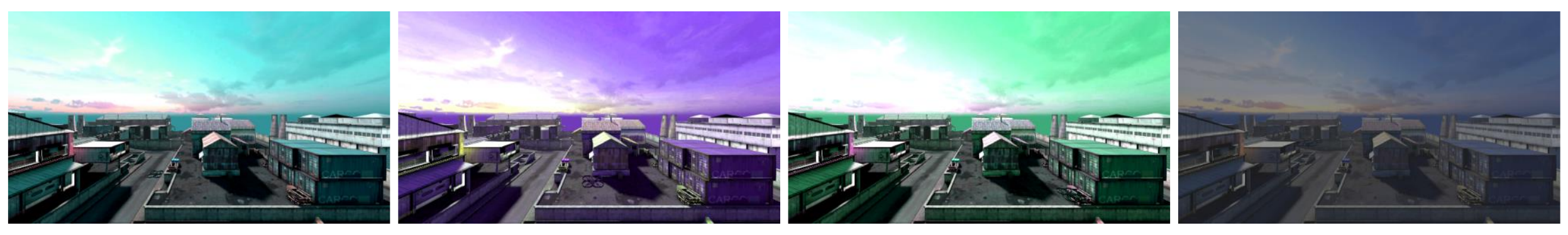}
\caption{Training set instances with domain randomization.}
\label{fig:dr}
\end{figure}

As displayed in Fig. \ref{fig:dr}, we vary the visual domain in collecting the training set. To reach better generalization, we regularize the pose estimation network using data augmentation. The used randomization factors \cite{muller2018driving} are:
\begin{itemize}
    \item Brightness: brightness factor is drawn from $(0.6,1.4)$
    \item Contrast: contrast factor is drawn from $(0.6, 1.4)$
    \item Saturation: saturation factor is chosen from $(0.6, 1.4)$
    \item Hue: hue factor is chosen from $(-0.2,0.2)$
\end{itemize}

\begin{figure}[b]
\centering
\includegraphics[width=.45\textwidth]{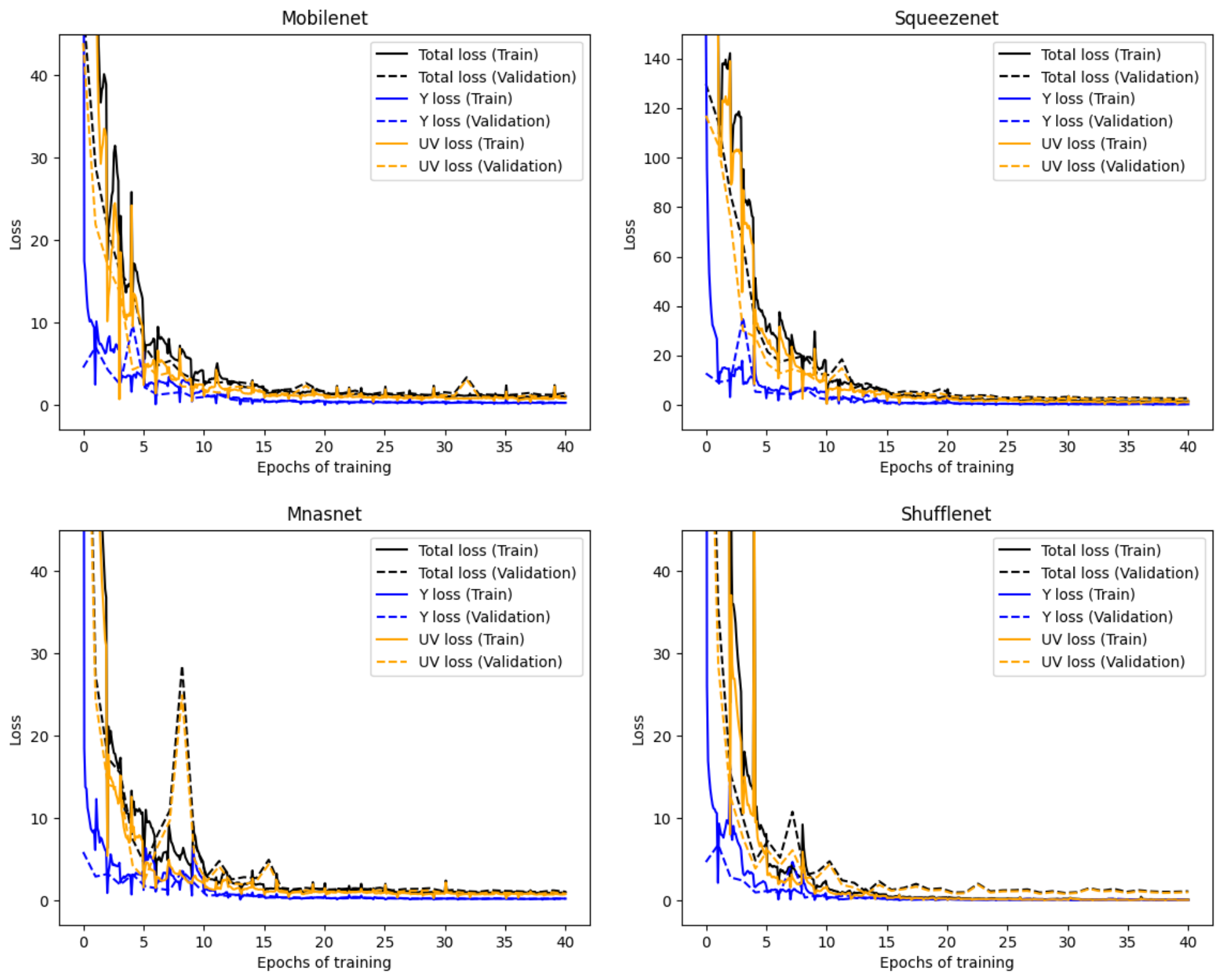}
\caption{Training curves for intruder pose estimation.}
\label{fig:training}
\end{figure}

\subsection{Training for intruder pose estimation \label{sec:training}}
We implemented all networks in Pytorch \cite{paszke2019pytorch} and used the Adam optimizer \cite{kingma2014adam}. A total of 4168 images were captured in the Unity scene with a dynamic defender. We set the initial learning rate to 5e-4, which was reduced by a factor of 2 every 5 epochs from epoch 10 until epoch 30. The final epoch of 40, a batch size of 4, and $\epsilon = 0.02$ for the loss function were used. 

We consider a number of backbones as possible alternatives to our model: MobileNet \cite{howard2017mobilenets}, SqueezeNet \cite{iandola2016squeezenet}, and MnasNet \cite{tan2019mnasnet}, and training curves of the four backbones are shown in Fig.~\ref{fig:training}. These are all lightweight networks that would enable real-time perimeter defense. We perform further training with domain randomization, which was applied to one-tenth of the training images, and split the training and validation sets with an $8:2$ ratio. We find that our model with the backbone as ShuffleNet \cite{zhang2018shufflenet} with domain randomization performs the best.

\begin{table}[t]
\centering
\caption{Mean Error for Pose Prediction}
\label{tab1}
\begin{center}
\begin{tabular}{c | c c}
\hline
Average Error & UV (pixel) & Y (m)\\
\hline
YOLOv3 \cite{redmon2018yolov3} & 13.94 & 2.81\\
MobileNet \cite{howard2017mobilenets}  & 11.06 & 0.52\\
SqueezeNet \cite{iandola2016squeezenet} & 11.39 & 0.66\\
MnasNet \cite{tan2019mnasnet} & 9.27 & 0.45\\
ShuffleNet \cite{zhang2018shufflenet} & 8.48 & 0.59\\
MobileNet + Domain randomization & 9.83 & 0.50\\
SqueezeNet + Domain randomization & 10.36 & 0.64\\
MnasNet + Domain randomization & 7.90 & \bf{0.44}\\
ShuffleNet + Domain randomization & \bf{6.25} & 0.46\\
\hline
\end{tabular}
\end{center}
\end{table}

The quantitative results for intruder pose prediction are detailed in Table \ref{tab1}. For comparison, we also run YOLOv3 detector \cite{redmon2018yolov3} that has been widely used for real-time object detection \cite{ramtoula2020msl, wang2021gdr, hong2021estimation}. Since YOLOv3 outputs a bounding box for detection, the UV prediction takes the center of the produced bounding box, and the $y_c$ prediction is estimated by looking at the size of the bounding box given that we assume the size of the intruder is known and fixed. In predicting the true pose of $(u,v)$, ShuffleNet with domain randomization achieves the lowest average UV error at 6.25 pixels. The $y_c$ prediction errors seem almost the same for all the methods (e.g. $0.55\pm 0.1m$), and this may be due to the fact that it is challenging to estimate the exact depth from an image since even a single pixel can represent a long distance. One exception is YOLOv3, which shows a much larger error in the $y_c$ prediction. This large error could be due to the fact that the size of the bounding box may not correctly represent the actual size of the intruder in case of occlusion.

\begin{figure}[b!]
\centering
\includegraphics[width=.4\textwidth]{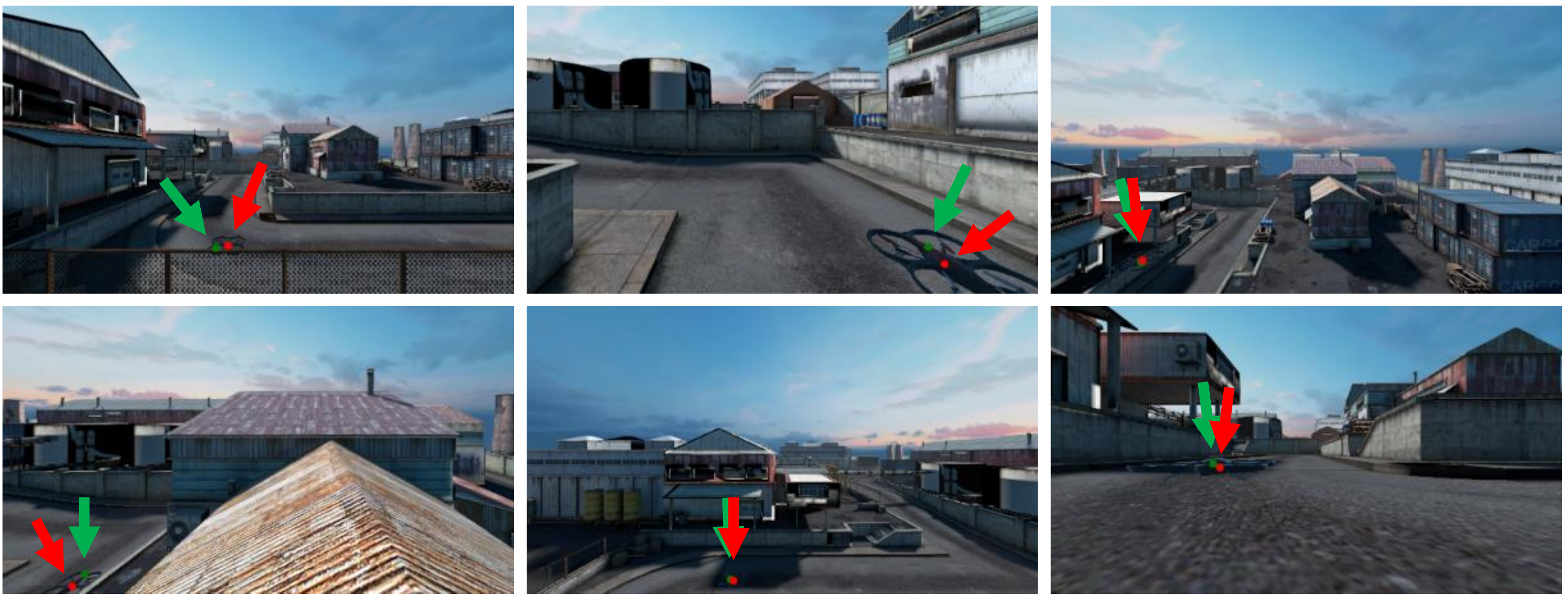}
\caption{True intruder pose (red) and prediction results (green) under occlusion, shadow, and different viewpoints.}
\label{fig:detection}
\end{figure}

\begin{figure}[t]
\centering
\includegraphics[width=.45\textwidth]{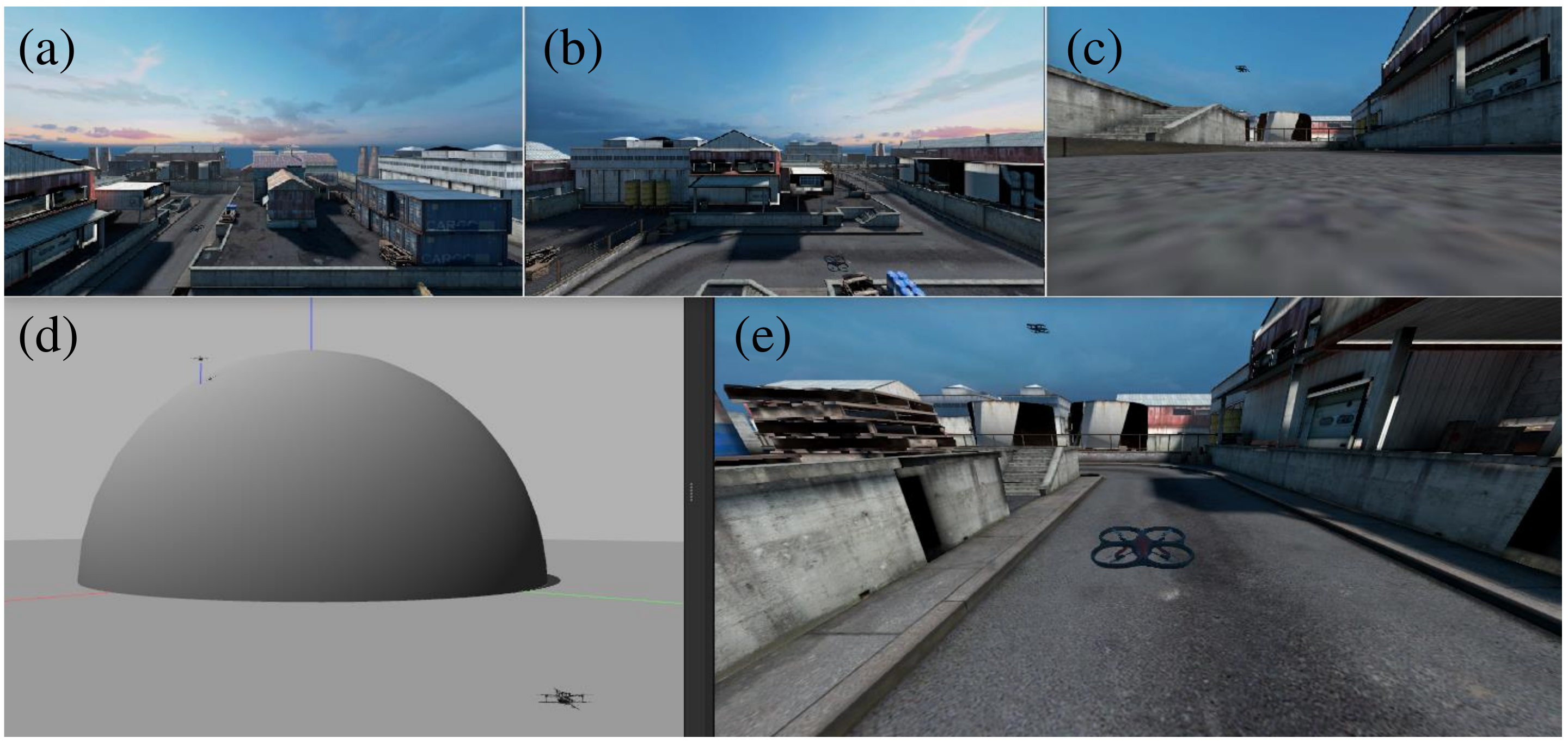}
\caption{Vision-based perimeter defense system. (a) Dynamic defender camera view. (b) Static defender camera view. (c) Intruder camera view. (d) Schematic view of agents and perimeter corresponding to the Flightmare simulator. (e) Overall Flightmare simulation environment.}
\label{fig:unity}
\end{figure}

Fig. \ref{fig:detection} shows the qualitative results for intruder pose prediction. The true and predicted intruder poses are marked by red and green circles, respectively. The pose estimator is robust against challenging environments considering occlusion, shadow, or different camera viewpoints.

\subsection{Ablation analysis on vision-based perimeter defense}
From the pose estimation results, we simulate the full defense system to close the perception-action loop. Fig. \ref{fig:unity} shows the overview of the vision-based perimeter defense. Fig. \ref{fig:unity}(a)-(c) are the camera views from the dynamic defender, static defender, and intruder. Fig. \ref{fig:unity}(d) shows the schematic view of agents and perimeter corresponding to the Flightmare simulator. Fig. \ref{fig:unity}(e) describes the overall Flightmare simulation \cite{song2020flightmare}. 

As shown in Fig. \ref{fig:overall}(a), the $(x_h,y_h,z_h)$ output from the multiview pose estimation is transmitted to the defender strategy module that computes the optimal breaching point based on agent positions. The control command for the defender is fed back into the Unity simulator to close the perception-action loop and render the moving agents.  

In the ablation analysis, we take an advantage of the static defender whose camera view has a different perspective from the dynamic defender's camera view. In particular, we assume that the dynamic defender's optical axis of its camera is perpendicular to the static defender's optical axis. 
Let $\{I_d,\overline{P}_d\}$ and $\{I_s,\overline{P}_s\}$ denote the true image and intruder pose pairs for the dynamic defender and static defender, respectively. Our goal is to learn the domain randomization mapping $f$ so that we can estimate the intruder pose given an image as $f(I_d) = \overline{P}_d$ and $f(I_s) = \overline{P}_s$.
The prediction $P_d$ and $P_s$ are the estimates of $\overline{P}_d$ and $\overline{P}_s$. We assume no errors in state estimation so the transformation $\mf T$ between the dynamic and static defenders are known (i.e. $\mf T (P_d) \simeq P_s$). Then, we can collect multiple data points $P_{d;k}$ from both defenders by
\begin{equation}
f(I_d) = P_{d;1} \text{  ,    }   \mf{T^{-1}}(f(I_s)) = P_{d;2} \;.
\end{equation}
While a previous work predicted the final pose as the sample average \cite{ren2019domain}, we combine $P_{d;1}$ and $P_{d;2}$ by parameterizing each camera coordinate as
\begin{equation}
x_c = x_{c;1}\alpha + x_{c;2}(1-\alpha),
\label{eq:xc}
\end{equation}
\begin{equation}
y_c = y_{c;1}\delta + y_{c;2}(1-\delta),
\label{eq:yc}
\end{equation}
\begin{equation}
z_c = z_{c;1}\gamma + z_{c;2}(1-\gamma),
\label{eq:zc}
\end{equation}
where $(x_{c;k},y_{c;k},z_{c;k})$ are the camera coordinates corresponding to $P_{d;k}$, and $\alpha, \delta, \gamma$ are parameters used for ablation study.

Table \ref{tab2} shows the relationship between the choice of parameters $\alpha, \delta, \gamma$ and performance metrics $\Delta d$. We set up the initial conditions in the experiments so that performance metrics should be zero if we assume the pose estimation was perfect. Trial 1 uses $\alpha=1.0, \delta=1.0, \gamma=1.0$, which represents the 1 vs. 1 defense where we only utilize the dynamic defender's perception module. Next, we give a high weight (e.g. 0.9) to a parameter in Trial 2 and give high weights to two parameters in Trial 3. Trial 4 shows various parameter choices with increasing weight on $\gamma$.

\begin{table}[t]
\centering
\caption{Ablation Analysis on Defense Algorithm}
\label{tab2}
\begin{center}
\begin{tabular}{c | c c c | c}
\hline
Trial & $\alpha$ & $\delta$ & $\gamma$ & $\Delta d$\\
\hline
1 & 1.0 & 1.0 & 1.0 & 4.91 \\
\hline
2(a) & 0.1 & 0.1 & 0.9 & 4.93 \\
2(b)  & 0.1 & 0.9 & 0.1 & 4.21 \\
2(c)  & 0.9 & 0.1 & 0.1 & \bf{4.14} \\
\hline
3(a)  & 0.1 & 0.9 & 0.9 & 4.44 \\
3(b)  & 0.9 & 0.1 & 0.9 & 4.38 \\
3(c)  & 0.9 & 0.9 & 0.1 & \bf{4.17} \\

\hline
\end{tabular}
\begin{tabular}{c | c c c | c}
\hline
Trial & $\alpha$ & $\delta$ & $\gamma$ & $\Delta d$ \\
\hline
4(a)  & 0.2 & 0.8 & 0.1 & 4.20 \\
4(b)  & 0.3 & 0.7 & 0.1 & 4.46 \\
4(c)  & 0.8 & 0.3 & 0.2 & 4.19 \\
4(d)  & 0.7 & 0.5 & 0.3 & 4.11 \\
4(e)  & 0.5 & 0.5 & 0.5 & 4.10 \\
4(f)  & 0.7 & 0.3 & 0.5 & \bf{4.03} \\
4(g)  & 0.9 & 0.1 & 0.5 & 4.15 \\
\hline
\end{tabular}
\end{center}
\end{table}

It can be inferred from Trials 2 and 3 that high weight on $\alpha$ would help minimize $\Delta d$. This indicates that it would be better to trust the dynamic defender than to trust the static defender in estimating the x-coordinate of the intruder in camera coordinates. This may be due to the fact that the dynamic defender estimates the x-coordinate as a lateral position while the static defender estimates it as longitudinal information. In general, the camera is known to obtain lateral information better than longitudinal information \cite{lee2017feature, lee2019bird}. It seems that the z-coordinate of the intruder is estimated the best if trusting both dynamic and static defenders with equal weight. 

In all trials, the choice of $\alpha=0.7, \delta=0.3, \gamma=0.5$ achieves the best performance, which results in $\Delta d=4.03$. It is worth noting that this result outperforms the choice of $\alpha=0.5, \delta=0.5, \gamma=0.5$, which represents the sample average method \cite{ren2019domain} in computing the final pose. As mentioned before, this ablation study validates that multiview perception improves the perimeter defense algorithm. 

\begin{figure}[b]
        \centering
        \subfloat[]{\includegraphics[width=4.2cm]{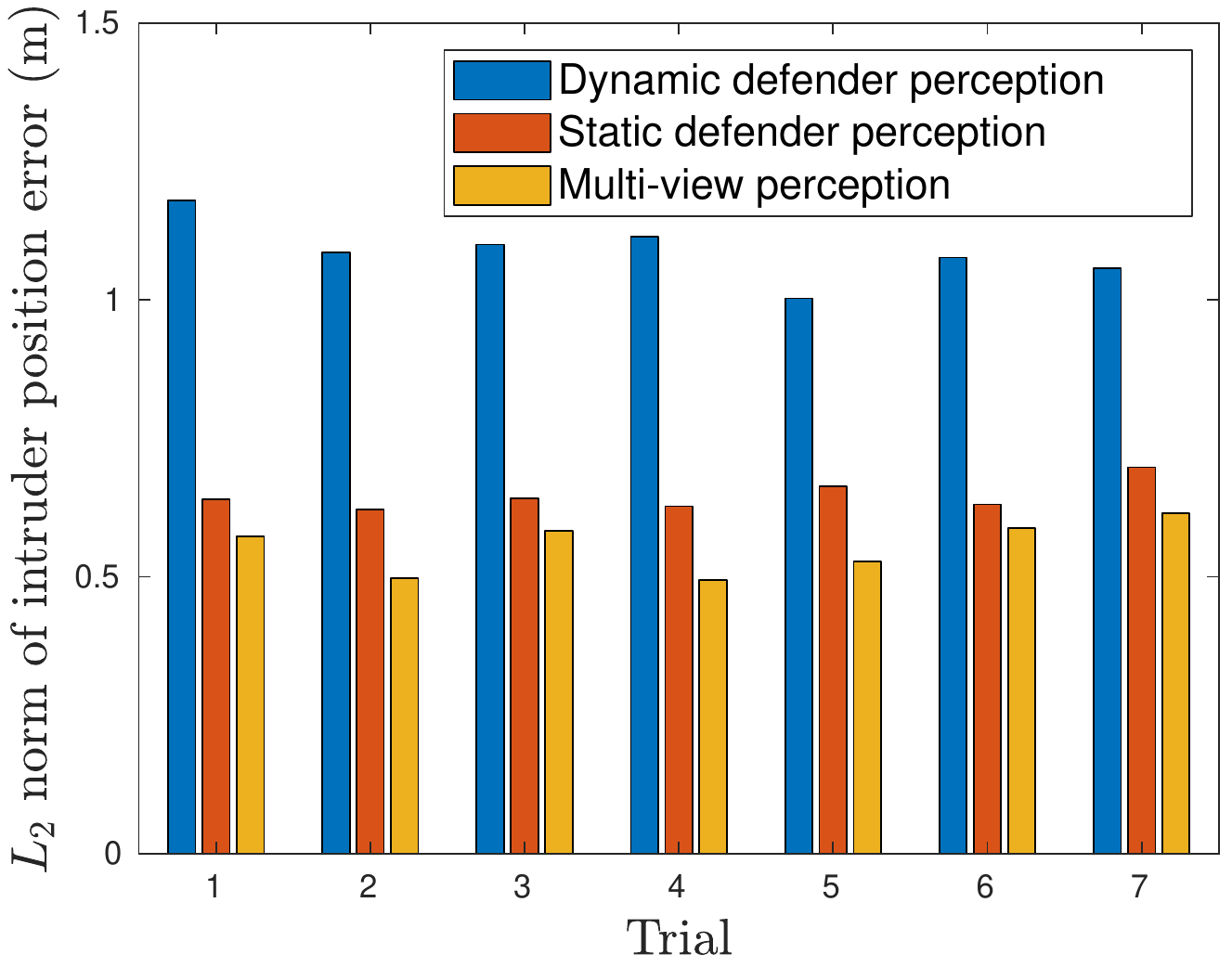}}
        \subfloat[]{\includegraphics[width=4.2cm]{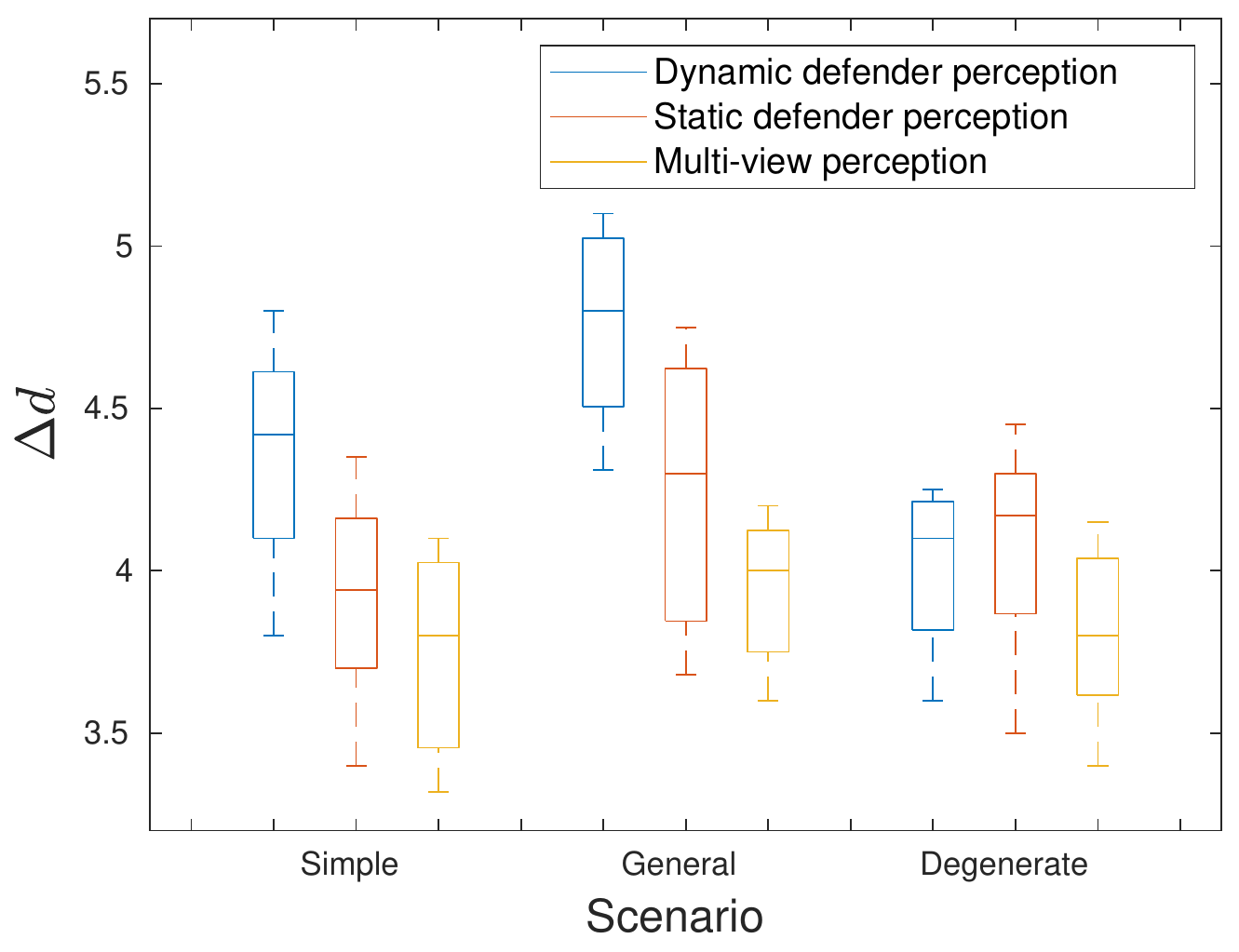}}
    \caption{(a) The average of $L_2$ norm of intruder position error with different perception methods. (b) Perimeter defense performance $\Delta d$ for representative scenarios.}
    \label{fig:graph}
\end{figure}

\subsection{Multiview perception on vision-based perimeter defense}
This section validates that multiview perception improves vision-based perimeter defense. Fig. \ref{fig:graph}(a) shows the intruder pose estimation error with three different choices for the perception component. For multiview perception, we use the parameters $\alpha=0.7,\delta=0.3,\gamma=0.5$ to achieve the best performance. In all cases, we assume that the intruder follows its optimal strategy so that we can equally compare a different set of defenses. We take the $L_2$ norm of intruder position error along the trajectories, and the average of $L_2$ norm position error in a trajectory is shown for a number of trials with different initialization. It can be seen that the static defender has less perception error than the dynamic defender does since the static defender is initially placed so that its perception can closely monitor the entire trajectory of the intruder. All trials verify that multiview perception lowers the $L_2$ norm of intruder position error from both defender viewpoints and thus improves the vision-based perimeter defense.

We evaluate the perimeter defense performance measured by $\Delta d$ using the three choices for the perception component, as shown in Fig. \ref{fig:graph}(b). We test the performance in representative scenarios: for "simple," both defender and intruder start with the same azimuth angles; for "general," defender moves on a geodesic path while intruder moves in a straight line towards the optimal breaching point; and for "degenerate," the initial position of the defender is on the base plane of the hemisphere. We run multiple trials with different initialization and confirm that multiview perception outperforms single-view perceptions in all scenarios. In the degenerate case, dynamic defender perception outperforms static defender perception since both agents are on the base plane.

\subsection{Hardware experiments}
\begin{figure}[b]
        \centering
        \subfloat[]{\includegraphics[width=3.5cm]{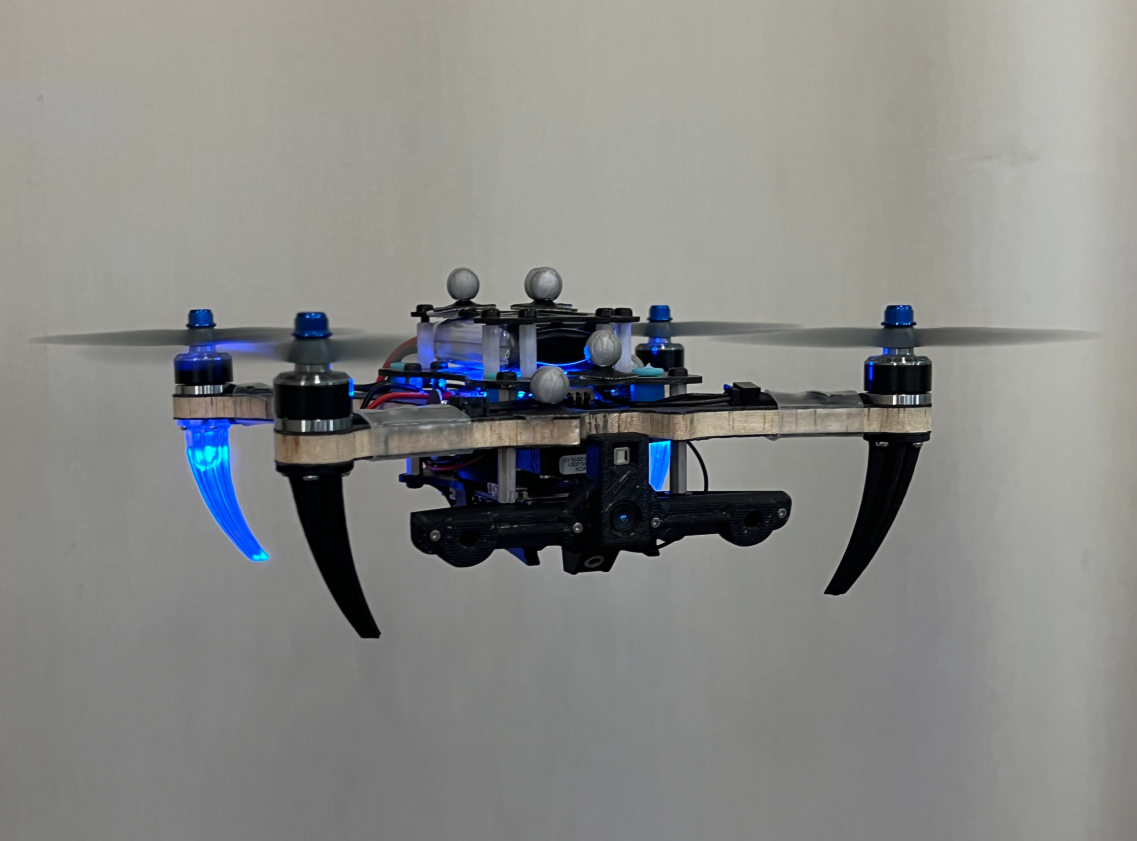}}
        \subfloat[]{\includegraphics[width=3.5cm]{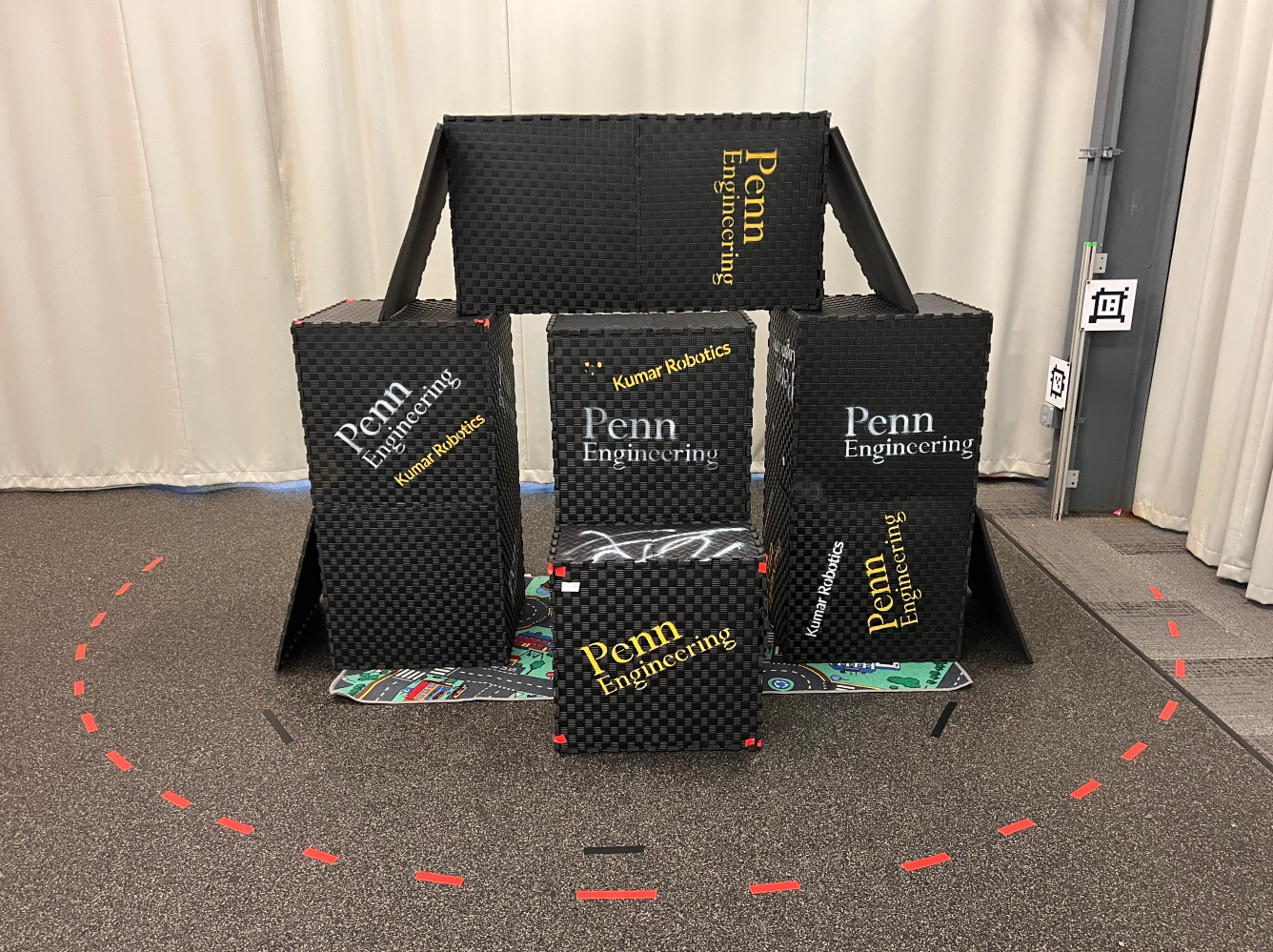}}
    \caption{Setup for hardware experiments. (a) The Snapdragon platform with a downward-facing camera. (b) Designed perimeter.}
    \label{fig:hardware}
\end{figure}

\begin{figure}[t]
        \centering
        \subfloat[]{\includegraphics[width=4.2cm]{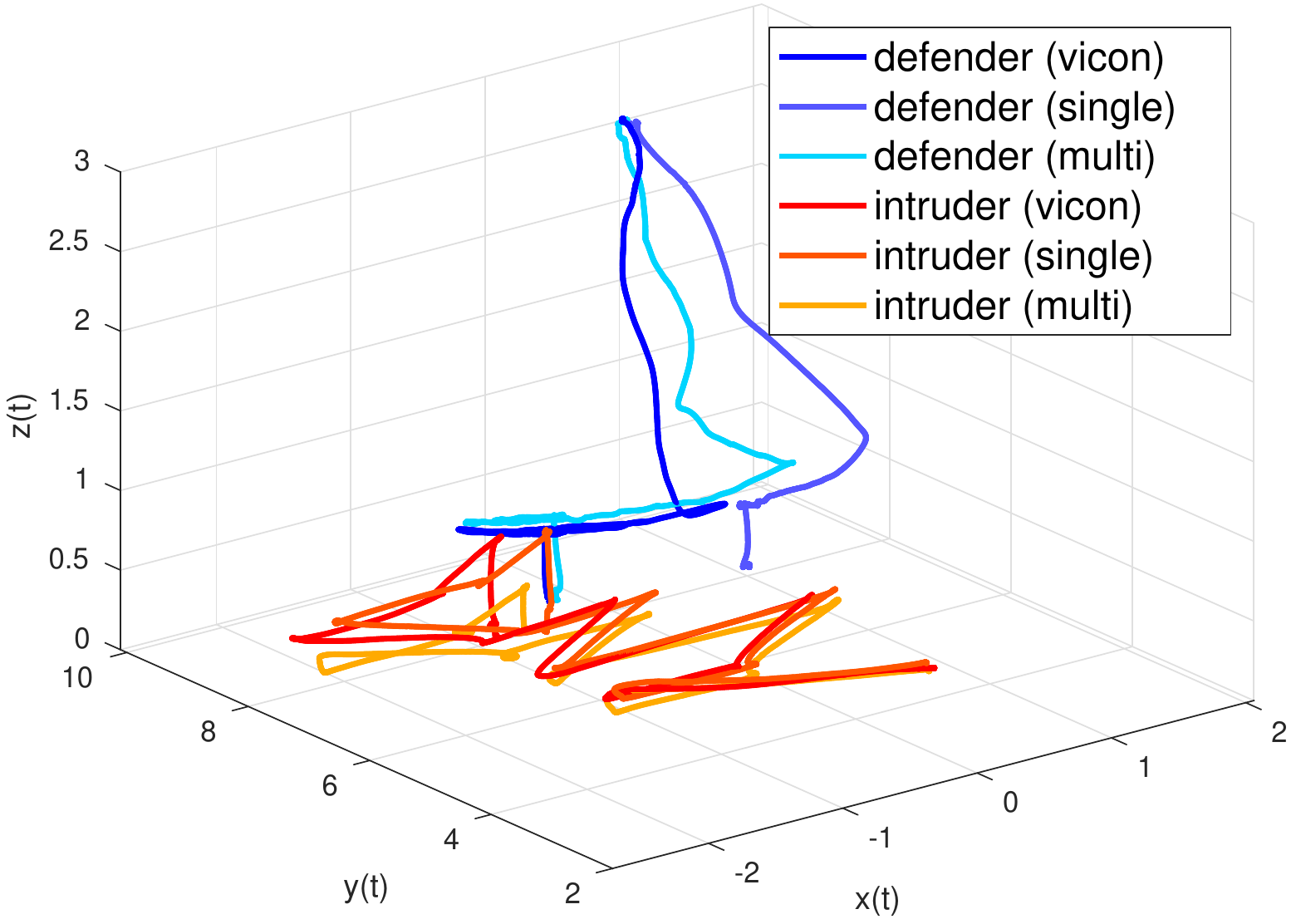}}
        \subfloat[]{\includegraphics[width=4.2cm]{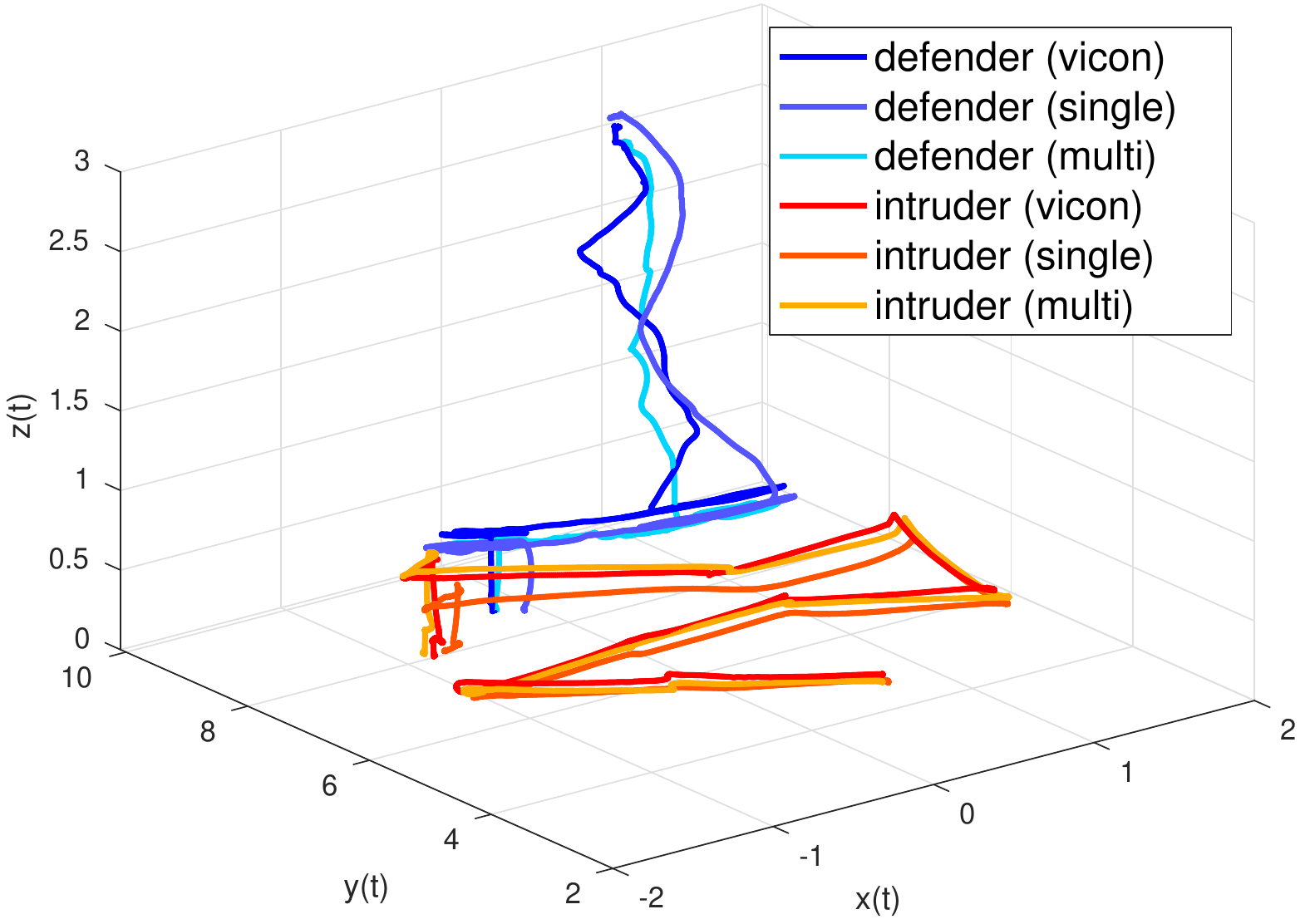}}
    \caption{(a)-(b) Defender and intruder trajectories for two perimeter defense games. For each game, we repeat a fixed set of attack trajectories for 1 vs. 1 (single) and 2 vs. 1 (multi) settings. We also run trials that defenders know the ground truth intruder poses for comparison (vicon).}
    \label{fig:traj}
\end{figure}

To further validate our proposed framework, we employ quadrotors as the defender and the intruder. The quadrotor platform is based on the quadrotor used in \cite{loianno2016estimation}. As shown in Fig.~\ref{fig:hardware}(a), it is a 0.16 m diameter, 256 g quadrotor using a Qualcomm Snapdragon$^{\text{TM}}$ board with a 7.4 V LiPo battery. The platform is equipped with a downward-facing camera and an on-board VIO state estimator. In the experiments, a dynamic defender hovers around a designed perimeter shown in Fig.~\ref{fig:hardware}(b) to prevent an intruder from reaching the perimeter, and a static defender on the side allows multiview perception (Fig.~\ref{fig:intro}). 

The robots' software stack is built on the Robot Operating System (ROS). All quadrotors run separate ROS masters and can communicate with the VICON Motion Capture system to obtain the ground truth pose. Due to the computing limit of the lightweight quadrotor platform, we perform the image processing on a base station but the image transmission and our full system are implemented in real-time. We utilize the same networks and parameters as the simulator to evaluate our proposed system.   

Fig.~\ref{fig:traj} shows the players' trajectories of two perimeter defense games. For each game, we repeat a fixed intruder trajectory and use three different state estimators for defenders to plan reactive trajectories. We use Vicon to plan the ground truth defender trajectory and compare single-view and multiview defense performance for 1 vs. 1 and 2 vs. 1 setting. We observe that in both games the trajectories planned by multiview perception are closer to the ground truth trajectories than those planned by single-view perception. It is worth noting in game 1 that the dynamic defender loses a track of the intruder when the intruder makes a sharp turn. In the 1 vs. 1 case, the defender stops following the intruder and stays still since the intruder is no longer within the field of view. However, in the 2 vs. 1 case, the dynamic defender keeps a track of the intruder with a help of the static defender. 

Table \ref{tab3} summarizes the average performance metrics $\Delta d$ over the entire trajectories played in each game. As can be seen in Fig.~\ref{fig:traj}, the noise level is higher in real-world experiments compared to the simulation case so taking an average throughout the game would provide a more meaningful interpretation. The $\Delta d$ values based on three state estimators (i.e., vicon, single-view, multiview) are all negatives, which indicates the winning of the defenders. The results show that by using multiview perception we obtain improved algorithm performances.

\begin{table}[t]
\centering
\caption{Defense Performance with State Estimators}
\label{tab3}
\begin{center}
\begin{tabular}{c | c c}
\hline
State Estimator & $\Delta d$ (avg. over game 1) & $\Delta d$ (avg. over game 2)\\
\hline
Vicon & -4.09 & -3.93\\
Single-view & -3.78 & -3.63\\
Multiview & -3.86 & -3.80\\

\hline
\end{tabular}
\end{center}
\end{table}
\section{Conclusion \label{sec:conclusion}}
This paper realizes vision-based perimeter defense and incorporates the pose estimation by aggregating multiple views. The multiview perception improves perimeter performance in simulation and real-world experiments. The future work will focus on working with multiple dynamic defenders with communication and consider an end-to-end perimeter defense approach.






%
%
%


\bibliography{reference}

\begin{thebibliography}{10}
\providecommand{\url}[1]{#1}
\csname url@rmstyle\endcsname
\providecommand{\newblock}{\relax}
\providecommand{\bibinfo}[2]{#2}
\providecommand\BIBentrySTDinterwordspacing{\spaceskip=0pt\relax}
\providecommand\BIBentryALTinterwordstretchfactor{4}
\providecommand\BIBentryALTinterwordspacing{\spaceskip=\fontdimen2\font plus
\BIBentryALTinterwordstretchfactor\fontdimen3\font minus
  \fontdimen4\font\relax}
\providecommand\BIBforeignlanguage[2]{{%
\expandafter\ifx\csname l@#1\endcsname\relax
\typeout{** WARNING: IEEEtran.bst: No hyphenation pattern has been}%
\typeout{** loaded for the language `#1'. Using the pattern for}%
\typeout{** the default language instead.}%
\else
\language=\csname l@#1\endcsname
\fi
#2}}

\bibitem{shishika2020review}
D.~Shishika and V.~Kumar, ``A review of multi agent perimeter defense games,''
  in \emph{International Conference on Decision and Game Theory for
  Security}.\hskip 1em plus 0.5em minus 0.4em\relax Springer, 2020, pp.
  472--485.

\bibitem{lee2020experimental}
E.~S. Lee, G.~Loianno, D.~Thakur, and V.~Kumar, ``Experimental evaluation and
  characterization of radioactive source effects on robot visual localization
  and mapping,'' \emph{IEEE Robotics and Automation Letters}, vol.~5, no.~2,
  pp. 3259--3266, 2020.

\bibitem{nguyen2019mavnet}
T.~Nguyen, S.~S. Shivakumar, I.~D. Miller, J.~Keller, E.~S. Lee, A.~Zhou,
  T.~{\"O}zaslan, G.~Loianno, J.~H. Harwood, J.~Wozencraft, \emph{et~al.},
  ``Mavnet: An effective semantic segmentation micro-network for mav-based
  tasks,'' \emph{IEEE Robotics and Automation Letters}, vol.~4, no.~4, pp.
  3908--3915, 2019.

\bibitem{chen2020sloam}
S.~W. Chen, G.~V. Nardari, E.~S. Lee, C.~Qu, X.~Liu, R.~A.~F. Romero, and
  V.~Kumar, ``Sloam: Semantic lidar odometry and mapping for forest
  inventory,'' \emph{IEEE Robotics and Automation Letters}, vol.~5, no.~2, pp.
  612--619, 2020.

\bibitem{lee2016drone}
S.~Lee, D.~Har, and D.~Kum, ``Drone-assisted disaster management: Finding
  victims via infrared camera and lidar sensor fusion,'' in \emph{2016 3rd
  Asia-Pacific World Congress on Computer Science and Engineering (APWC on
  CSE)}.\hskip 1em plus 0.5em minus 0.4em\relax IEEE, 2016, pp. 84--89.

\bibitem{lee2020perimeter}
E.~S. Lee, D.~Shishika, and V.~Kumar, ``Perimeter-defense game between aerial
  defender and ground intruder,'' in \emph{2020 59th IEEE Conference on
  Decision and Control (CDC)}.\hskip 1em plus 0.5em minus 0.4em\relax IEEE,
  2020, pp. 1530--1536.

\bibitem{shishika2020cooperative}
D.~Shishika, J.~Paulos, and V.~Kumar, ``Cooperative team strategies for
  multi-player perimeter-defense games,'' \emph{IEEE Robotics and Automation
  Letters}, vol.~5, no.~2, pp. 2738--2745, 2020.

\bibitem{yan2019construction}
R.~Yan, Z.~Shi, and Y.~Zhong, ``Construction of the barrier for reach-avoid
  differential games in three-dimensional space with four equal-speed
  players,'' in \emph{2019 IEEE 58th Conference on Decision and Control
  (CDC)}.\hskip 1em plus 0.5em minus 0.4em\relax IEEE, 2019, pp. 4067--4072.

\bibitem{tobin2017domain}
J.~Tobin, R.~Fong, A.~Ray, J.~Schneider, W.~Zaremba, and P.~Abbeel, ``Domain
  randomization for transferring deep neural networks from simulation to the
  real world,'' in \emph{2017 IEEE/RSJ international conference on intelligent
  robots and systems (IROS)}.\hskip 1em plus 0.5em minus 0.4em\relax IEEE,
  2017, pp. 23--30.

\bibitem{vidal2002probabilistic}
R.~Vidal, O.~Shakernia, H.~J. Kim, D.~H. Shim, and S.~Sastry, ``Probabilistic
  pursuit-evasion games: theory, implementation, and experimental evaluation,''
  \emph{IEEE transactions on robotics and automation}, vol.~18, no.~5, pp.
  662--669, 2002.

\bibitem{arola2019uav}
S.~Arola and M.~A. Akhloufi, ``Uav pursuit-evasion using deep learning and
  search area proposal,'' in \emph{2019 IEEE international conference on
  robotics and automation}.\hskip 1em plus 0.5em minus 0.4em\relax IEEE, 2019.

\bibitem{labbe2020cosypose}
Y.~Labb{\'e}, J.~Carpentier, M.~Aubry, and J.~Sivic, ``Cosypose: Consistent
  multi-view multi-object 6d pose estimation,'' in \emph{European Conference on
  Computer Vision}.\hskip 1em plus 0.5em minus 0.4em\relax Springer, 2020, pp.
  574--591.

\bibitem{ramtoula2020msl}
B.~Ramtoula, A.~Caccavale, G.~Beltrame, and M.~Schwager, ``Msl-raptor: A 6dof
  relative pose tracker for onboard robotic perception,'' \emph{arXiv preprint
  arXiv:2012.09264}, 2020.

\bibitem{wang2021gdr}
G.~Wang, F.~Manhardt, F.~Tombari, and X.~Ji, ``Gdr-net: Geometry-guided direct
  regression network for monocular 6d object pose estimation,'' in
  \emph{Proceedings of the IEEE/CVF Conference on Computer Vision and Pattern
  Recognition}, 2021, pp. 16\,611--16\,621.

\bibitem{ren2019domain}
X.~Ren, J.~Luo, E.~Solowjow, J.~A. Ojea, A.~Gupta, A.~Tamar, and P.~Abbeel,
  ``Domain randomization for active pose estimation,'' in \emph{2019
  International Conference on Robotics and Automation (ICRA)}.\hskip 1em plus
  0.5em minus 0.4em\relax IEEE, 2019, pp. 7228--7234.

\bibitem{lee2021defending}
E.~S. Lee, D.~Shishika, G.~Loianno, and V.~Kumar, ``Defending a perimeter from
  a ground intruder using an aerial defender: Theory and practice,'' in
  \emph{2021 IEEE International Symposium on Safety, Security, and Rescue
  Robotics (SSRR)}.\hskip 1em plus 0.5em minus 0.4em\relax IEEE, 2021, pp.
  184--189.

\bibitem{zhang2018shufflenet}
X.~Zhang, X.~Zhou, M.~Lin, and J.~Sun, ``Shufflenet: An extremely efficient
  convolutional neural network for mobile devices,'' in \emph{Proceedings of
  the IEEE conference on computer vision and pattern recognition}, 2018, pp.
  6848--6856.

\bibitem{mellinger2011minimum}
D.~Mellinger and V.~Kumar, ``Minimum snap trajectory generation and control for
  quadrotors,'' in \emph{2011 IEEE international conference on robotics and
  automation}.\hskip 1em plus 0.5em minus 0.4em\relax IEEE, 2011, pp.
  2520--2525.

\bibitem{song2020flightmare}
Y.~Song, S.~Naji, E.~Kaufmann, A.~Loquercio, and D.~Scaramuzza, ``Flightmare: A
  flexible quadrotor simulator,'' 2020.

\bibitem{muller2018driving}
M.~M{\"u}ller, A.~Dosovitskiy, B.~Ghanem, and V.~Koltun, ``Driving policy
  transfer via modularity and abstraction,'' \emph{arXiv preprint
  arXiv:1804.09364}, 2018.

\bibitem{loquercio2019deep}
A.~Loquercio, E.~Kaufmann, R.~Ranftl, A.~Dosovitskiy, V.~Koltun, and
  D.~Scaramuzza, ``Deep drone racing: From simulation to reality with domain
  randomization,'' \emph{IEEE Transactions on Robotics}, vol.~36, no.~1, pp.
  1--14, 2019.

\bibitem{paszke2019pytorch}
A.~Paszke, S.~Gross, F.~Massa, A.~Lerer, J.~Bradbury, G.~Chanan, T.~Killeen,
  Z.~Lin, N.~Gimelshein, L.~Antiga, \emph{et~al.}, ``Pytorch: An imperative
  style, high-performance deep learning library,'' \emph{arXiv preprint
  arXiv:1912.01703}, 2019.

\bibitem{kingma2014adam}
D.~P. Kingma and J.~Ba, ``Adam: A method for stochastic optimization,''
  \emph{arXiv preprint arXiv:1412.6980}, 2014.

\bibitem{howard2017mobilenets}
A.~G. Howard, M.~Zhu, B.~Chen, D.~Kalenichenko, W.~Wang, T.~Weyand,
  M.~Andreetto, and H.~Adam, ``Mobilenets: Efficient convolutional neural
  networks for mobile vision applications,'' \emph{arXiv preprint
  arXiv:1704.04861}, 2017.

\bibitem{iandola2016squeezenet}
F.~N. Iandola, S.~Han, M.~W. Moskewicz, K.~Ashraf, W.~J. Dally, and K.~Keutzer,
  ``Squeezenet: Alexnet-level accuracy with 50x fewer parameters and< 0.5 mb
  model size,'' \emph{arXiv preprint arXiv:1602.07360}, 2016.

\bibitem{tan2019mnasnet}
M.~Tan, B.~Chen, R.~Pang, V.~Vasudevan, M.~Sandler, A.~Howard, and Q.~V. Le,
  ``Mnasnet: Platform-aware neural architecture search for mobile,'' in
  \emph{Proceedings of the IEEE/CVF Conference on Computer Vision and Pattern
  Recognition}, 2019, pp. 2820--2828.

\bibitem{redmon2018yolov3}
J.~Redmon and A.~Farhadi, ``Yolov3: An incremental improvement,'' \emph{arXiv
  preprint arXiv:1804.02767}, 2018.

\bibitem{hong2021estimation}
Y.~Hong, J.~Liu, Z.~Jahangir, S.~He, and Q.~Zhang, ``Estimation of 6d object
  pose using a 2d bounding box,'' \emph{Sensors}, vol.~21, no.~9, p. 2939,
  2021.

\bibitem{lee2017feature}
E.~S. Lee and D.~Kum, ``Feature-based lateral position estimation of
  surrounding vehicles using stereo vision,'' in \emph{2017 IEEE Intelligent
  Vehicles Symposium (IV)}.\hskip 1em plus 0.5em minus 0.4em\relax IEEE, 2017,
  pp. 779--784.

\bibitem{lee2019bird}
E.~S. Lee, W.~Choi, and D.~Kum, ``Bird’s eye view localization of surrounding
  vehicles: Longitudinal and lateral distance estimation with partial
  appearance,'' \emph{Robotics and Autonomous Systems}, vol. 112, pp. 178--189,
  2019.

\bibitem{loianno2016estimation}
G.~Loianno, C.~Brunner, G.~McGrath, and V.~Kumar, ``Estimation, control, and
  planning for aggressive flight with a small quadrotor with a single camera
  and imu,'' \emph{IEEE Robotics and Automation Letters}, vol.~2, no.~2, pp.
  404--411, 2016.

\end{thebibliography}
\bibliographystyle{IEEEtran}

\end{document}